\pdfoutput=1

\documentclass[11pt]{article}

\usepackage{emnlp2021}

\usepackage{times}
\usepackage{latexsym}

\usepackage[T1]{fontenc}

\usepackage[utf8]{inputenc}

\usepackage{microtype}
\usepackage{multirow}
\usepackage{adjustbox}
\usepackage{graphicx}
\usepackage{amssymb}
\usepackage{setspace}
\usepackage{comment}
\usepackage{amsmath}
\usepackage[ruled, vlined]{algorithm2e}
\usepackage{algorithmicx}
\usepackage{algpseudocode}
\title{A Simple Yet Efficient Method for Adversarial Word-Substitute Attack}

\author{Tianle Li \and Yi Yang \\
  The Hong Kong University of Science and Technology \\
  \texttt{tliax@connect.ust.hk, imyiyang@ust.hk}}

\begin{document}
\maketitle
\begin{abstract}
NLP researchers propose different word-substitute black-box attacks that can fool text classification models. In such attack, an adversary keeps sending crafted adversarial queries to the target model until it can successfully achieve the intended outcome. State-of-the-art attack methods usually require hundreds or thousands of queries to find one adversarial example. In this paper, we study whether a sophisticated adversary can attack the system with much less queries. We propose a simple yet efficient method that can reduce the average number of adversarial queries by 3-30 times and maintain the attack effectiveness. This research highlights that an adversary can fool a deep NLP model with much less cost.
\end{abstract}

\section{Introduction}

Deep learning models are vulnerable to adversarial examples \cite{szegedy2013intriguing,goodfellow2014explaining,papernot2016transferability}. A burgeoning literature in NLP community studies the word-substitute based black-box adversarial attack \citep{alzantot2018generating,jin2019bert,vijayaraghavan2019generating,alzantot2019genattack,morris2020textattack}. In the black-box attack setting, an adversary has no access to the target NLP model (such as a sentiment classifier) architecture, parameters or training data, but he is capable of querying the target model with crafted inputs and obtaining the output predictions. By querying the target model, the adversary is able to exploit the model weakness and therefore find the adversarial examples. 
An example of adversarial attack is present in Table \ref{table:example}. 


While it seems that the adversarial examples impose practical threat to the NLP systems, it is interesting to note that a state-of-the-art method usually needs to send hundreds or even thousands of crafted examples to the target model, in order to generate \textbf{one} successful adversarial example \citep{alzantot2018generating,jin2019bert}. Is it really a practical threat any more? First, if the adversary attacks a cloud-based text classification system \citep{papernot2017practical}, it may cost the adversary about \$0.5 - \$1 to generate one adversarial example \footnote{{Take Google Cloud Platform for example,  assume 1,000 queries (sentences) are used to generate one adversarial example, and average query contains 500 characters. That would be equivalent to 500 ``units'' in Google API which costs \$0.5 -\$1.  \url{https://cloud.google.com/natural-language/pricing} }}. Obviously, in practice, an adversary is budget constrained. Second, if the adversary attacks a commercial in-house NLP system, it would also be suspicious (since it is like a DDOS attack) to the host given the large number of queries sent.

\begin{table*}[t]
\begin{adjustbox}{width=1\textwidth}
\begin{tabular}{|l|l|l|}
\hline
\multicolumn{3}{|l|}{\textbf{Original Text Prediction: Positive}}

                                                                                        \\ \hline
\multicolumn{3}{|l|}{\multirow{2}{*}{\begin{tabular}[c]{@{}l@{}}The individuals at FastMed are \textcolor{red}{wonderful}, I made an appointment using there APPT app got in \textcolor{red}{quickly} thanks\\ for caring. \end{tabular}}} \\
\multicolumn{3}{|l|}{}                                                                                                                                                                                                                                                                                                 \\ \hline
\multicolumn{3}{|l|}{\textbf{Adversarial Text Prediction: Negative}}                                                                                                                                                                                                                                                   \\ \hline
\multicolumn{3}{|l|}{\begin{tabular}[c]{@{}l@{}}The individuals at FastMed are \textcolor{blue}{wondrous}, I made an appointment using there APPT app got in \textcolor{blue}{sooner} thanks\\ for caring. \end{tabular}}  
                                                                                            \\ \hline
\end{tabular}
\end{adjustbox}
\caption{One example and its adversarial example generated by our approach.}
\label{table:example}
\end{table*}

Can a sophisticated adversary attack the system with much fewer queries? 
Intuitively, the unsuccessful  queries, although failed, may still contain critical information that can help the adversary to better design the subsequent queries. After accumulating more and more successful and unsuccessful queries, the adversary may therefore greatly reduce the total number of queries needed.
In this work, we propose a method that reduces the queries, regardless of which adversarial method is used. 
We build a simple logistic regression model using all the intermediate queries in the generation of previous adversarial examples in order to determine which words are more likely to lead to a successful attack. We also simultaneously narrow the range of search space by  defining a dynamically updated rule according to the real-time attack process.

In the experiments, we combine our approach with greedy-search based \citep{jin2019bert} and genetic-search based attacks \citep{alzantot2018generating}. The results show that the average number of queries can be reduced by 3-30 times, while maintaining the same level of attack effectiveness. 
While the method proposed in this paper is very simple, it provides a new angle to the adversarial attack. This research highlights that an adversary can launch an attack on NLP systems with much less cost.

\section{Related Work}

Adversarial attack on NLP systems has drawn great attention from NLP researchers and practitioners. See \citep{zhang2020adversarial} for an overview. Given the discrete nature of human language, attacks based on word-substitute, character-substitute, word-deletion, etc. \cite{gao2018black,ren2019pwws,Li_2019} have been proposed to generate adversarial examples in text with undetectable perturbations from human beings. Recent literature has focused on word-substitute black-box attack \citep{jin2019bert,alzantot2018generating,alzantot2019genattack,li2020bert,garg2020bae}. In a nutshell, the word-substitute attack searches for adversarial examples by determining the words to replace first and then the candidates of the selected words. The Greedy-search algorithm \citep{jin2019bert,garg2020bae,li2020bert} or genetic-search algorithm \citep{alzantot2018generating,alzantot2019genattack} is commonly adopted in the search procedure. 



While the prior works mostly focus on developing different search algorithms for adversarial example generation, one thing in common is that they have ignored the previous unsuccessful queries. Our work differs from the prior works as we examine whether an adversary can utilize those unsuccessful queries to design the subsequent attack. 

\section{Method}

A word substitute black-box attack usually contains two steps. First, an adversary selects a target word to substitute. For example, as shown in Table \ref{table:example}, \textit{wonderful} is the selected target word. This step is known as \textbf{Word Ranking}.
Second, the adversary selects a synonym to replace the target word. In the example, \textit{wondrous} is the chosen word to replace \textit{wonderful}. This step is known as \textbf{Word Replacement}.
The adversary keeps sending queries to the target model (such as a sentiment classifier) until a successful adversarial is found, i.e., the prediction outcome is altered. This procedure involves a \textit{combinatorial search} of word perturbations, where greedy-search or genetic-search algorithm is commonly used.


Intuitively, even though the intermediate queries fail to attack the model, they may still contain critical information that assist the adversary in the subsequent adversarial generation. Therefore, we propose two strategies to improve the word ranking and word replacement steps respectively.



\noindent\textbf{{Problem Formulation}}
An adversary aims to attack a target model $G$ with a set of document examples $D = \{d_1,...,d_n \}$, using existing adversarial attack method $F$. It takes $F$ number of $q_i$ queries to find the corresponding adversarial example $d_i^{adv}$ for the $i$-th document $d_i$. The adversarial examples would alter the predictions of the target model so that $G(d_i^{adv}) \neq G(d_i)$. 
Our goal is to minimize the average number of queries needed, i.e., $\sum_i (q_i)/{n}$.


\subsection{Word Ranking Strategy}
 The prior methods enumerate over words in a sentence and obtain the word importance scores. For example, \citep{jin2019bert} defines the word importance scores as the prediction change before and after deleting the word in a sentence. Although straightforward, this operation is expensive as it sends a number of queries that is equivalent to the sentence length. 

We propose that the adversary can assess the word importance from  the successful/unsuccessful queries. In other words, the adversary can estimate the likelihood that a word is an important ``adversarial'' word. Prior research states that the adversarial examples exist due to the presence of low-probability region in the manifold \citep{szegedy2013intriguing}. Therefore, if certain words are highly correlated with the previously successful examples, it is possible that those words are near the low-probability region, where the target model is the most vulnerable. 

In particular, we train a simple discriminative model whose inputs are the previous queries and the outputs are the corresponding indicators of the attacking results with respect to the target model $G$.  At the beginning of the attack, when the adversary has no query yet, he can select important words using the default word ranking method \citep{jin2019bert}. With the attacks going on, the adversary obtains a good number of queries $Q_{inter}$ together with their labels $Y_{inter}$, indicating whether they can successfully attack the target model or not. Since we need to obtain the importance score of each word feature, we use logistic regression $LogiReg$: $Q_{inter} \rightarrow Y_{inter}$. We deploy GloVe-200 \citep{pennington-etal-2014-glove} to encode each word in $Q_{inter}$. The coefficients of word features can be interpreted as the degree of the contribution a specific word makes to the consequence whether a given example could be a successful adversarial example. Note that under this setting, word importance is no longer measured as its contribution to the prediction outcome. 

\SetArgSty{textnormal}
\begin{algorithm}[t]
\setstretch{0.3}
\SetAlgoLined
\SetKwInOut{Input}{Input}
\SetKwInOut{Output}{Output}
\Input{Documents $D = \{d_1,...d_n\}$; Target model $G$; Attack method $F$.}
\Output{The corresponding adversarial examples $D^{adv} = \{d_1^{adv},...d_n^{adv}\}$}
initialization\ $Q = \emptyset$\;
 \For{$d_i \in D$}{
    \If{Q not $\emptyset$}{
    Rank word $w \in d_{i}$ by its $\beta_{w}$
    }
    \While{True}{
    Word Ranking: select target word $w$ to perturb;\\
    Word Replacement: select synonym $w^\prime$ to replace;
    generate query $q_j = F(d_i, w \rightarrow w^\prime)$ and obtain prediction outcome $y_j = G(q_j)$;\\
        add $(q_j$, $I(y_j = G(d_i))$ in $Q$.\\ 
        if $y_j \neq G(d_i)$, ${d_i}^{adv} = q_j$, adversarial found, exit.
    } 
    Train logistic regression with $Q$ and obtain word coefficient ${\beta}_w$;
 }
 \caption{Proposed model-agnostic word-substitute attack.}
\end{algorithm}

\subsection{Word Replacement Strategy}
Another vital task of generating adversarial examples in text is to select the synonyms to replace the chosen target words. As shown in Table \ref{table:example}, the word ``wondrous'' is selected to replace the target word ``wonderful''. The criterion is to choose a word which can flip the prediction outcome (thus adversarial) while maintaining a semantic similarity with the original example so that it is imperceptible to human beings.


The pretrained word embeddings are commonly employed to choose the nearest synonym neighbors of a selected word \citep{jin2019bert, alzantot2018generating, ebrahimi2018hotflip}. Although it is a simple and direct method, it is also expensive as it needs to query a large number of synonyms. Therefore, a natural question is, can an adversary choose the synonyms more efficiently? To employ the previous queries within a single example, we proposed a simple strategy to choose the synonyms based on the moving direction of the embedding vectors of the words. The intuition is that if the substitution of a word can decrease the confidence score to a large degree, it can implicate the approximate direction in the embedding space that drives the sentence nearer to the decision boundary of the target model, which can be utilized in the following synonyms search and decrease the number of queries.

In particular, we first choose the $N$ nearest neighbors based on the cosine similarity of counter-fitting vectors \cite{mrk2016counterfitting}, which demonstrates the superb capability of judging the similarities among words. After the replacement of any word decreases the confidence score of the target model $G$, we can compute the direction vector from the original word to its substitution in the counter-fitting vector space. 
The direction vector that achieves the largest decrease in the confidence score can be set for the moving direction of the next selected word. Therefore, in the next replacement, merely $K$ synonyms with the top moving direction match in terms of cosine similarity are chosen from the $N$ nearest neighbors. And we query the target model $G$ with each of the $K$ replacement and maintain the change with the largest drop of confidence score for the original prediction.
The complete adversarial procedure is present in Algorithm 1.

\begin{table}[htb]
\centering
\begin{adjustbox}{width=0.45\textwidth}
\begin{tabular}{c|cccc}
\hline
Dataset   & \# Classes & Avg Len & \begin{tabular}[c]{@{}c@{}}\# Examples \\ (train)\end{tabular} & \begin{tabular}[c]{@{}c@{}}\# Examples \\ (attack)\end{tabular} \\ \hline
IMDB      & 2          & 215     & 25K                                                            & 1K                                                            \\
Yelp      & 2          & 152     & 560K                                                           & 1K                                                            \\
AG's News & 4          & 43      & 120K                                                           & 1K                                                            \\ \hline
\end{tabular}
\end{adjustbox}
\caption{Dataset description.}
\label{table:4}
\end{table}

\begin{table}[htb]
\small
\begin{tabular}{c|c|ccc}
\hline
               &  Method                             & Acc\% & Per\% & \#Q \\ \hline
\multirow{4}{*}{IMDB} &  Genetic                            & 45.7          & 4.9   & 6493    \\
                      &                       Genetic + Ours                     & 13.6               & 7.6       & 212         \\
                      &                        Greedy                             & 13.6               & 6.1       & 1134        \\
                      &                       \multicolumn{1}{l|}{Greedy + Ours} & 13.4               & 6.1       & 404         \\ \hline
\multirow{4}{*}{Yelp} &  Genetic                            & 31.0           & 10.1 & 6137    \\
                      &                       Genetic + Ours                     & 9.4                & 12.1      & 187         \\
                      &                       Greedy                             & 6.6                & 13.9      & 827         \\
                      &                       \multicolumn{1}{l|}{Greedy + Ours} & 2.7                & 13.7      & 471         \\ \hline
\multirow{4}{*}{AG}   & Genetic                            & 51.0           & 16.9 & 3495    \\
                      &                       Genetic + Ours                     & 21.5               & 20.1      & 177         \\
                      &                        Greedy                             & 12.5               & 22.0      & 357         \\
                      &                       \multicolumn{1}{l|}{Greedy + Ours} & 12.2               & 26.6      & 188         \\ \hline
\end{tabular}
\caption{Experiment results of original baselines and after employing our algorithm. The original accuracy of the target model is 87.8\% (IMDB), 95.6\% (Yelp) and 94.2\% (AG) respectively. Genetic results are those reported in \citep{li2020bert}. Greedy results are obtained using authors' released code \citep{jin2019bert}. Acc, Per and \#Q short for after-attack accuracy, perturbation rate and number of queries respectively.}
\label{table:main}
\end{table}

\begin{table}[]
\setlength{\belowcaptionskip}{-0.5cm}
\small
\begin{tabular}{c|c|ccc}
\hline
 & Method &  Acc\% & Per\% & \#Q \\ \hline
\multirow{4}{*}{IMDB} & Greedy                      & 13.6  & 6.1 & 1134    \\
                      & Greedy+WRankS               & 12.1   & 8.7  & 430         \\
                      & Greedy+WRepS                & 11.2   & 4.2 & 829         \\
                      & Greedy+Both                 & 13.4 & 6.1 & 404         \\ \hline
\end{tabular}
\caption{Ablation study.}
\label{tab:ablation}
\end{table}

\vspace{-0.5cm}
\section{Evaluation}

Word-substitute attack can be mostly grouped into two categories: \textbf{genetic}-search based \citep{alzantot2019genattack} and \textbf{greedy}-search based \citep{jin2019bert}. In the experiments, we evaluate the proposed strategies with both of the methods. 

We report three metrics: \textbf{Attack Accuracy} measures the prediction accuracy of the target model on the adversarial examples, \textbf{Perturb} measures the average percentage of words being changed, \textbf{Avg Queries} denotes the average number of queries needed for one example. For an attack method, low attack accuracy, low change rate and low average number of queries are desirable for the adversary.

Following the prior literature, we use three datasets: IMDB review dataset, Yelp review dataset and AG news dataset in the experiments. The datasets details are present in Table \ref{table:4}. For each dataset, $1000$ randomly selected sentences from test set are used for adversarial attack.
The target models used in the experiments are the pretrained BERT models that are fine tuned on three datasets respectively, as released by \citep{jin2019bert}.

The main results are present in Table \ref{table:main}. Our method is model-agnostic and can be incorporated with either greedy-based or genetic-based word substitute adversarial attack. The results show that our method can significantly reduce the number of average queries of both greedy or genetic method by 3-30 times.

For genetic algorithm, we implemented a variation of the genetic attacking strategy in \cite{alzantot2018generating}. We replaced the original method of calculating fitness score for individual word with the word ranking strategy we proposed, which is set to be the probabilities of being selected as the altered words in the first generation. With $\textit{Maximum Generation} \approx 140$, $\textit{Population} \approx 5$, $N \approx 30$ and $K \approx 5$, the average number of queries required to generate an adversarial examples is decreased from $6493$ to $212$ on IMDB dataset with significantly lowered after-attack accuracy and slightly higher perturbation rate. And the same level of improvements are achieved on Yelp and AG datasets in Table \ref{table:main}. For greedy algorithm, we substituted the important words ranking policy and synonym candidates selection strategy accordingly based on the proposed methods and preserved the other settings the same as the previous work \citep{jin2019bert}. As shown in Table \ref{table:main}, we decrease approximate half of the average number of queries while maintaining the same level of adversarial accuracy and perturbation rates as previous methods after setting $N \approx 100$ and $K \approx 30$. 
An example of adversarial example generated by our approach is also present in Table \ref{table:example}. 

\vspace{-0.0cm}
We conduct ablation studies on the the effectiveness of word ranking strategy, word replacement strategy and both strategies combined. The result is present in Table \ref{tab:ablation}. Due to the space limit, we show the performance of using greedy-search \citep{jin2019bert} as the backend attack method on IMDB dataset. The result shows that both of the strategies can decrease the average number of queries on the greedy-based method. And combining both strategies together leads to a further reduction on the number of queries. The results are consistent on the other datasets with the genetic-based method as well. 

%


\vspace{-0.1cm}
\section*{Conclusion}

In this work, we present a simple yet efficient method for adversarial word-substitute attack. The core idea of our method is to leverage the successful and failed adversarial queries to guide the efficient search for word substitutions. The experiment results show that an adversary can reduce the query numbers substantially. The proposed method is also model-agnostic and it can be incorporated with the existing greedy or genetic based attacks. We hope this work can provide a different angle in adversarial attack and contribute to the growing literature and practice on NLP system safety.

\section*{Ethics/Broader Impact Statement}

The work has potential to make contribution to adversarial NLP research and practice. While the state of the art adversarial NLP methods usually need hundreds or thousands of queries to generate one adversarial example, we show that a sophisticated adversary can greatly reduce the query numbers by exploiting the information from the previous successful/failed queries. Our approach is simple, easy to implement, and model-agnostic that can be incorporated into either greedy-based or genetic-based word-substitute attack approaches. This research further raises the awareness of NLP system safety.

\bibliography{anthology,custom}
\bibliographystyle{acl_natbib}

\end{document}